\documentclass{article}

\usepackage{arxiv}

\usepackage[utf8]{inputenc} 
\usepackage[T1]{fontenc}    
\usepackage{hyperref}       
\usepackage{url}            
\usepackage{booktabs}       
\usepackage{amsfonts}       
\usepackage{nicefrac}       
\usepackage{microtype}      
\usepackage{lipsum}		
\usepackage{graphicx}
\usepackage{natbib}
\usepackage{multirow}
\usepackage{doi}
\usepackage{amsmath}

\usepackage{pifont}
\usepackage{placeins}

\newcommand{\cmark}{\ding{51}}%
\newcommand{\xmark}{\ding{55}}%

\usepackage[dvipsnames]{xcolor}
\definecolor{cornflowerblue}{RGB}{100,149,237}
\definecolor{alizarin}{rgb}{0.82, 0.1, 0.26}
\newcommand{\elicit}{ELICIT}

\usepackage{enumitem}
\usepackage{cleveref}  

\hypersetup{
    colorlinks,
    linkcolor={cornflowerblue},
    citecolor={cornflowerblue},
    urlcolor={alizarin}
}

\begin{document}
\title{Optimising Human-Machine Collaboration for Efficient High-Precision Information Extraction from Text Documents}

\date{}

\author{Bradley Butcher\thanks{Authors contributed equally to this research.} \\
  University of Sussex \\
  The Alan Turing Institute\\ 
  UK \\
   \\\And
  Miri Zilka\footnotemark[1] \\
  University of Cambridge \\ Cambridge, UK \\
  \texttt{mz477@cam.ac.uk} \\\And
  Darren Cook \\
  University of Liverpool \\ Liverpool, UK \\
\And
  Jiri Hron \\
  University of Cambridge \\ Cambridge, UK \\
  \And
  Adrian Weller \\
  University of Cambridge \\ The Alan Turing Institute \\ UK \\
  }
  
%




\renewcommand{\shorttitle}{Optimising Human-Machine Collaboration for Information Extraction}

\maketitle
\begin{abstract} \looseness=-1
While humans can extract information
from unstructured text
with high precision and recall, this is often too time-consuming to be practical. 
Automated approaches, on the other hand, produce nearly-immediate results, but may not be reliable enough for high-stakes applications where precision is essential. 
In this work, we consider the benefits and drawbacks of 
various human-only, human-machine, and machine-only information extraction approaches. 
We argue for the utility of a human-in-the-loop approach in applications where high precision is required, but purely manual extraction is infeasible.
We present a framework and an accompanying tool for information extraction using weak-supervision labeling with human validation.
We demonstrate our approach on three criminal justice datasets. 
We find that the combination of the computer speed and human understanding yields precision comparable to manual annotation while requiring only a fraction of time, and significantly outperforms fully automated baselines in terms of precision.
\end{abstract}



\keywords{Human-Computer Collaboration, Human-in-the-loop, Information Extraction, Weak Supervision}


\maketitle

\section{Introduction}
\label{sec:introduction}

\looseness=-1
High precision extraction of structured 
data from long texts has valuable applications in 
medicine \citep{wang2018clinical}, finance \citep{ding2015deep}, science \citep{nasar2018information, weston2019named, olivetti2020data}, and law enforcement \citep{waltl2018rule, carnaz2018automated, hung2019recognizing}. 
While humans often achieve near-perfect retrieval precision and recall, their work is time-intensive, costly, and they can be traumatised by the content (e.g., sex, violence) \cite{criddle_2021}. 
On the other hand, even state-of-the-art automated methods \citep{grishman2019twenty,zaman2020information} still cannot extract information from text {with high precision} \citep{grishman2019twenty}.

\looseness=-1
We investigate tools for structured information extraction (IE) with emphasis on the \emph{trade-off between processing speed and precision}.
To ground our research, we focus on use cases within the criminal justice domain.
The data sources in this area---whether publicly available \citep{ormachea2015new,angwin2016machine,Rudin2020Age,justfair} or restricted \citep{justice_2022}---are usually derived from routinely collected administrative data.
Such data often omits key information like victim details, or mitigating and aggravating circumstances \citep{Zilka2022datasets}.
This information, and much more, can however be found in long unstructured documents like trial transcripts.
Making such information easily accessible would enable much needed research into racial and gender bias in the judicial system.

\looseness=-1
We focus on three distinct datasets and two IE tasks.
Two datasets contain information about criminal cases from legal documents and news articles. 
For both, the task is to extract structured (tabular) information about the victims, defendants, and situational circumstances. 
The third dataset contains online communications between convicted child sexual predators, and police officers posing as minors; the task is to identify specific predatory behavior patterns \citep{Cook2022}. 
In all cases, the text documents are too long for \emph{manual} extraction at scale \citep{Jiang2021}, 
and the high precision necessary for the extracted data to be useful to researchers
rules out fully \emph{automated} approaches.
We therefore focus on human-computer collaboration.

\looseness=-1
Settings where computers are quicker but less accurate than humans are under-explored in the 
literature \citep{rastogi2022unifying, assael2022restoring}. 
In \Cref{sec:setting}, we discuss how this type of human-computer complementarity affects 
design choices within collaborative IE,
and identify a gap among existing solutions (see \Cref{tab:tool_overview}).
In \Cref{sec:elicit}, we fill this gap by developing \elicit{}, a new human-validated IE tool which leverages the speed of modern large language models, and the near-perfect human accuracy.
We build on weak supervision approaches \citep{ratner2017snorkel, zhang2022survey}, utilising several automated methods to provide suggestions from which the user selects the final answer.
We achieve precision on-par with manual annotation---significantly outperforming the state-of-the-art among \emph{automated} IE tools \citep{ratner2017snorkel, zhang2022survey}---while using orders of magnitude less processing time (\Cref{sec:results}).
In \Cref{sec:results_adv}, we further explore: (i)~increasing the number of computer-generated suggestions to improve recall; (ii)~deferral as a way to trade-off performance for time; and (iii)~ranking and fine-tuning as auxiliary ways to use the validated data to attain even better performance.
Our main contributions are summarized below:
\begin{enumerate}
    \item 
    We analyze the relative strengths of various human-computer approaches to IE from long unstructured texts.
    
    \item 
    We design and implement \elicit{}, an interactive IE tool which combines weak supervision and human validation. 
    
    \item 
    Creating three new datasets, we demonstrate ELICIT significantly speeds up the annotation process, while maintaining near-perfect precision.
    Accuracy can be further traded-off for processing time via a deferral setting.
    
\end{enumerate}

\section{Requirements of high-precision information extraction}
\label{sec:setting}

\looseness=-1
IE and data labeling are used in a variety of settings, each with a distinct set of challenges.
Not all settings are similarly suitable for human-computer collaboration \citep{Jiang2021}, and practitioners often resist solutions which over focus on automation instead of providing assistance \citep{Feuston2022}. 
To ground our investigation of human-computer collaboration in IE,
we start by exploring two use cases related to the criminal justice domain (\Cref{sec:use_cases}).
Centering the discussion around use cases allows us to examine how attributes of different tasks affect system desiderata.
This enables us to evaluate and compare available solutions, and identify existing gaps in the literature and tool landscape (\Cref{sec:existing_tools}).

\subsection{Criminal justice use cases}
\label{sec:use_cases}

\subsubsection{Task~1: Criminal trial information}
\label{sec:task_crime}

Research in criminal justice is severely hindered by lack of high-quality publicly available datasets \citep{lammy2017lammy, millar2011whitewashing, logan2016policing, douglas2017risk}.
A great deal of useful information is contained in unstructured documents such as court transcripts, sentencing remarks, judgements, and press articles. 
Extracting this information into a structured format can enable research into critical issues like racial bias in criminal justice \citep{kovera2019racial,clemons2014blind}.
The documents are however too numerous and lengthy for manual extraction.
This creates a need for a \emph{time-efficient} yet \emph{accurate} IE tool.
To enable high-quality extraction across a wide variety of settings, the system should also be \emph{flexible}, and able to take advantage of efficient IE methods for a given task when they exist.

To ensure \emph{reproducibility} of any analysis facilitated by the tool, it should be feasible to document the dataset creation steps in a way that allows independent reconstruction.
This is a particular concern under ambiguity.
To illustrate, consider extracting whether a victim or suspect was legally considered vulnerable from sentencing remarks.\footnote{ \looseness=-1
Sentencing remarks summarize the judge's ruling in a criminal trial. 
They typically describe the crime, and any mitigating or aggravating circumstances.}
Such information is seldom mentioned explicitly---unless directly relevant to the ruling---but can sometimes be inferred.
While humans are skilled at making such inferences, they often need to read a large portion of the text for context, and may disagree with each other's judgments. 
These disagreements, and a lack of explicit rules for resolving them, are a core problem in reproducibility \citep{obels2020analysis}. 
When adding any human element, it is thus key to establish guidelines around how potentially ambiguous data should be recorded.

\subsubsection{Task~2: Online child sexual exploitation discourse}
\label{sec:task_grooming}

Online sexual grooming of minors is an increasing problem in the digital age \citep{greene2020experiences}.
Many abusers seek to establish physical contact offline \citep{shelton2016online}. 
To assist the law enforcement, 
academics have been trying to in advance identifying predators who steer the relationship towards physical encounters \citep{briggs2011exploratory, winters2017sexual, o2003typology, williams2013identifying}. 
To date, this important work has relied on time-consuming manual annotation of conversations involving child sex offenders. 
While full automation is possible, it currently comes at a cost of low accuracy \citep{Cook2022}.

\looseness=-1
Our second task is identifying signs of offline contact solicitation in online chats.
The goal is to annotate conversations in accordance with the `self-regulation' theory of child grooming \citep{elliott2017self}, which has already been employed in \citep{Cook2022}.
In practice, this means detecting whether each conversational instance (i.e., a continuous message exchange without more than a one-hour break) contains any of the following \emph{offender behaviors}: 
(1)~rapport building, 
(2)~control, 
(3)~challenges, 
(4)~negotiation, 
(5)~use of emotions, 
(6)~testing boundaries, 
(7)~use of sexual topics, 
(8)~mitigation, 
(9)~encouragement,
(10)~risk management.
See \citep{Cook2022} for a qualitative description of these behaviors, and further discussion.

\looseness=-1

When done manually, an expert annotator scans the offender's messages one-by-one, deciding for each whether one or more of the above ten behaviors is present.
For reference, performing manual annotation on 24 chats took a forensic psychologist over 600 hours \citep{Cook2022}.
Designing a solution with high \emph{time-efficiency} and \emph{accuracy} is therefore crucial. 
Reducing the user's exposure to this difficult content is a significant additional benefit \cite{steiger2021psychological}.
For optimal results, the system should again be \emph{flexible} enough to enable incorporation of already existing IE solutions.
Finally, due to the more subjective nature, \emph{reproducibility}, or the ability easily compare disagreement in annotations is even more pertinent here than for Task~1 (\Cref{sec:task_crime}). 
In particular, since annotators can substantially disagree, we want the data creation process to be reproducible at the level of these differences. We can then monitor and see if there is significant inter-annotator disagreement.
Since we require the tool to be time-efficient, using multiple human annotators is significantly more feasible than with fully manual annotation. 

\subsubsection{Comparing the tasks}
\label{sec:task_comparison}

General design considerations of human-in-the-loop IE tools based on existing workflows have been studied in \citep{Rahman2022}.
However, the specificity of our use-cases (\Cref{sec:task_crime,sec:task_grooming}) requires that we explore the design space in context of our goals \citep{mackeprang2019discovering}. 
Inspecting the previous sections,
we identified several \textbf{common requirements}:
\begin{itemize}
    \item \looseness=-1
    \textbf{Time-efficiency}, i.e., 
    orders of magnitude faster than manual extraction.
    
    \item
    \textbf{Accuracy}, i.e.,
    the results contain little incorrect information (\emph{precision}), and are as complete as possible (\emph{recall}).
    
    \item
    \textbf{Flexibility}, i.e., 
    can extract variety of user-specified information, and incorporate existing IE tools.
    
    \item 
    \textbf{Reproducibility}, i.e.,
    the information extraction process should be replicable with appropriate documentation, and inter-annotator disagreement should be trackable.

    \item 
    \textbf{Shielding}, i.e., reducing the user's exposure to the text by only highlighting relevant sections. 
\end{itemize}
Time-efficiency is essential due to the large quantities of text;
due to the high-stakes nature of the criminal justice domain, we particularly 
prioritise that the speed up results in almost no precision loss of the extracted information.

\subsection{Existing tools and their drawbacks}
\label{sec:existing_tools}

\begin{table}[]
\caption{\label{tab:tool_overview} Different human-computer configurations applicable to IE (\Cref{sec:existing_tools}), and the degree to which they meet the requirements for our use cases as identified in \Cref{sec:task_comparison}.}
\begin{tabular}{l|cccccc}

& flexibility & precision & recall & reproducibility & time-efficiency  & shielding \\
\midrule
human-only & +++ & +++ & +++ & depends & \text{-} \text{-} \text{-} & \text{-} \text{-} \text{-} \\
human + search & +  & +++ & \multicolumn{2}{l}{\hspace{0.9em} --- trade-off ---} & \multirow{5}{*}{\includegraphics[width=0.03\textwidth, height=18.5mm]{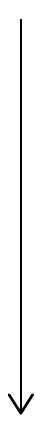}} & - \\
assisted annotation & + & +++ & + & depends & &  \multirow{4}{*}{\includegraphics[width=0.03\textwidth, height=13.9mm]{figs/arrow.pdf}} \\
validation (top-$K$) & depends & +++ & + & + & & \\
deferral & + & \multirow{2}{*}{\includegraphics[width=0.03\textwidth, height=8mm]{figs/arrow.pdf}} & +  & depends & \\
human feedback for training & depends  &  & + & + &  & \\
fully-automated & +  & + & + & +++  & +++ & +++                       
\end{tabular}
\end{table}


We now discuss various approaches for addressing the two tasks from \Cref{sec:task_crime,sec:task_grooming} with focus on their key requirements (see \Cref{tab:tool_overview} for an overview).

\subsubsection{Manual extraction \& search}
\label{sec:human_search}

\looseness=-1
Humans with appropriate background are the gold standard for both tasks (given sufficient time and motivation).
However, ensuring high recall requires the reader to scan through the whole text, which is prohibitively slow for our use cases \citep{Jiang2021}.
Besides fatigue, both our tasks (\Cref{sec:use_cases}) expose the reader to potentially disturbing content which may impact their well-being \cite{arsht2018human}.
Humans can take advantage of contextual knowledge, and make flexible on-the-fly decisions.
While often advantageous, this flexibility may hinder reproducibility when no fixed protocol is followed, or when the task requires a level of subjective judgment (as in Task~2).

Human labeling can be sped up by making the source documents searchable. 
Beyond digitisation, this may take the form of keyword search, or regular expression matching.
An example of a relevant tool is OpenSearch \citep{opensearch}, an open-source fork of the more well-known Elasticsearch \citep{elasticsearch}.
While this preserves flexibility and accuracy, the boost to time-efficiency is often small, especially when assigning a correct label requires understanding large portions of the text (as in Task~2).
Reproducibility concerns also remain, unless a very specific set of rules is detailed and followed.
If such rules exist, an approach with a greater level of automation may produce similar performance more efficiently.

\subsubsection{Assisted annotation}
\label{sec:assist_annotation}

\looseness=-1
Assisted annotation tools are commonly used across many domains, including law, medicine, and political science \citep{neves2014survey,stoykov2019legal, zadgaonkar2021overview, haddadan2019yes}.
Their aim is to produce annotated text, with labels assigned to each annotation.
The tools are most often used in an iterative process: 
the algorithm proposes annotations, the user makes modifications to correct mistakes, the tool uses this feedback to learn better recommendations, and so on.
For example, \citep{Desmond2022} uses a semi-supervised model to predict the most probable labels for each instance. They show this increases speed and accuracy of data labeling.
Some other recent examples of these tools are: prodi.gy, lighttag, and CLIEL \citep{garcia2017cliel, Prodigy, perry2021lighttag}.
Since the user has full control over the final annotation, accuracy is ensured provided there is sufficient time and motivation.
The algorithmic proposals can provide a speed-up, but it is limited in cases where the user is still required to read or scan through large chunks of the document.
For the same reason, the reduction of user fatigue, and exposure to disturbing material, is rather limited.
Analogously to \Cref{sec:human_search}, reproducibility can be difficult to achieve.



\subsubsection{Human validation}
\label{sec:human_validation}

\looseness=-1
By human validation, we mean methods where automated algorithms make all label predictions, \emph{each} of which is then reviewed by a human. 
When combined with \emph{passage retrieval} \citep[e.g.,][]{karpukhin2020dense},
this setup exploits the complementary of human and computer abilities (accuracy and speed).
Beyond time-efficiency and accuracy, human validation also improves reproducibility, at least at the level of the machine predictions. 
The level inter-annotator agreement can be measured by using multiple annotators and improved, if needed, by providing clear guidelines.
The only open-source tool from this category we found is Rubrix \citep{rubrix}.
Rubrix satisfies many of our desiderata.
However, its main purpose is labeling of many short texts, rather than extracting a set of interdependent variables from larger documents.

\subsubsection{Deferral to human validator} 
\label{sec:deferral}

\looseness=-1
Deferral is an \emph{extension} of human validation (\Cref{sec:human_validation}) where the human reviews \emph{only some} of the predictions.
This allows trading-off accuracy for time-efficiency, and alleviates user fatigue and exposure to harmful material.
The cases to defer are typically chosen using an estimate of prediction confidence.
When these estimates are well-calibrated, 
significant time savings may be attained with little performance loss.
Deferral can be particularly useful for large amounts of relatively low-stake decisions, e.g., moderating social media comments \citep{Lai2022}.
However, calibrating confidence estimates remains a challenge, especially in deep learning \citep{guo2017calibration}.
Without calibration, potential for improvement is often marginal.
Deferral style solutions can be obtained by combining
any Human validation (\Cref{sec:human_validation})
and automated labeling (\Cref{sec:human_training,sec:full_automation}) algorithm, provided the latter outputs confidence estimates.

\subsubsection{Human validation in training}
\label{sec:human_training}

Validation using human experts can be costly and time-consuming.
An alternative to deferring to humans during labeling is to only use their feedback for model training.
We specifically refer to a form of active learning where the user is only asked to label several strategically selected examples at the beginning, after which the fine-tuned model labels the rest of the dataset in a fully automatic mode.
An example of a tool from this class is IWS \citep{boecking2021interactive}.
Compared to human validation (\Cref{sec:human_validation}) and deferral (\Cref{sec:deferral}), this method enables further time-savings at the price of performance reduction. 
While the impact on reproducibility and user fatigue \& well-being is positive, the drop in performance is often too large to satisfy the requirement of near-perfect precision (\Cref{sec:task_comparison}).

\subsubsection{Full automation}
\label{sec:full_automation}

\looseness=-1
Automated IE is an area of active research \citep{adnan2019limitations,grishman2019twenty}.
Algorithms in this category do not defer any of their predictions to humans.
This provides the best time-efficiency and reproducibility. 
However, even state-of-the-art algorithms \citep{ratner2017snorkel,ratner2019training,grishman2019twenty,zaman2020information, layoutlm} cannot achieve the level of performance our use cases require.

\begin{figure}[t]
    \centering
    \includegraphics[width=\textwidth]{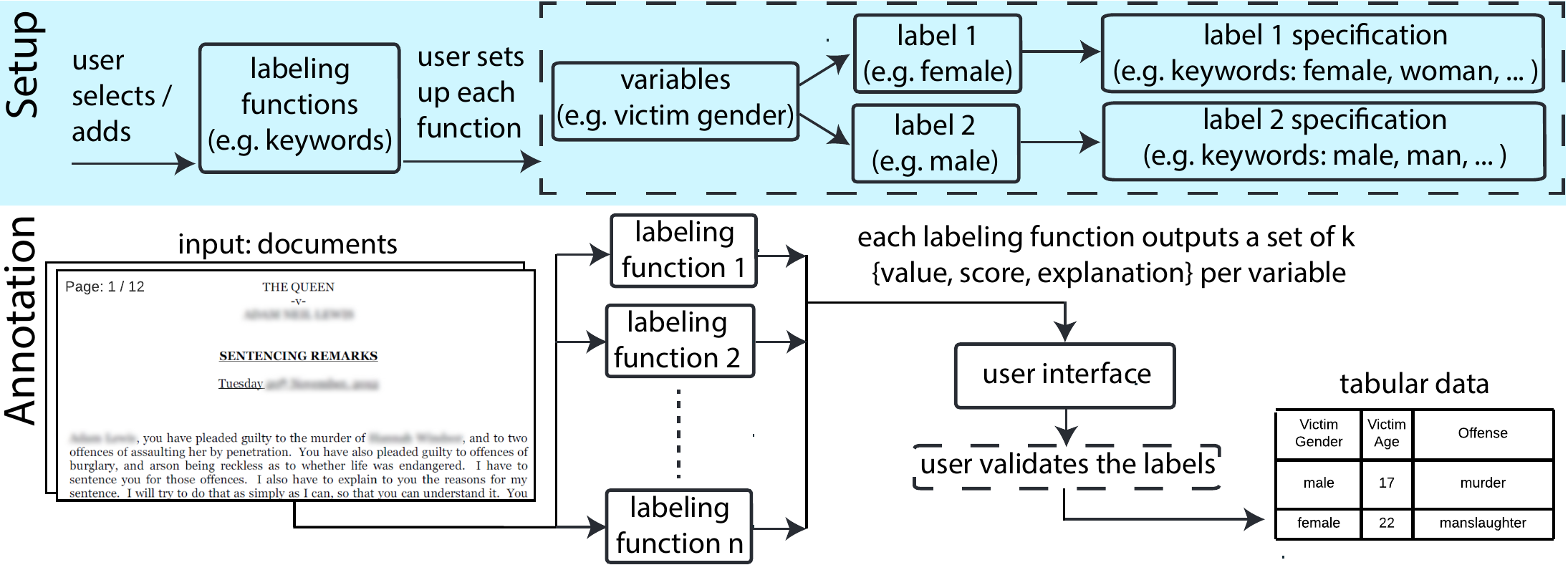}
    \caption{\looseness=-1
    \textbf{A high-level diagram of the \elicit{} framework.} \emph{Setup:} User chooses the labeling functions, and the information (variables) to be extracted;
    the labeling functions 1) identify relevant sections of the text; and 2) suggest a label for the user to validate.
    \emph{Annotation:}
    For each variable, labeling functions provide possible values and accompanying explanations. 
    The user validates these, creating a tabular dataset where each row corresponds to a document, and each column to an extracted variable.
    }
    \label{fig:overview}
\end{figure}

\section{\elicit{}: A system for user-validated information extraction from text}
\label{sec:elicit}

\subsection{System design}

\looseness=-1
Based on the \Cref{sec:task_comparison} requirements, we designed and implemented \elicit{}, a flexible, accurate, and time-efficient tool for IE from text.
Its core functionality falls into the \emph{human validation} category (\Cref{sec:human_validation}), with specialization on extraction of interdependent variables from long complex texts.
\elicit{} consists of two high-level stages (\Cref{fig:overview}): 
(1)~automated passage retrieval and label suggestion, and 
(2)~human label validation. 
The prediction \emph{accuracy} relies on the user's ability to infer the correct label from the retrieved passages, while the \emph{time-efficiency} gains come mainly from the automated passage retrieval. 
Since the user only interacts with text excerpts, they are \emph{shielded} from having to engage with the entirety of the potentially disturbing text.
The automated passage retrieval and label suggestions also allows for step~(1) to be \emph{reproducible}, and step~(2) easily comparable between annotators.

\looseness=-1
To ensure \emph{flexibility}, the automated first step utilizes weak supervision---a method for combining multiple {`weak labels'} generated by \textit{labeling functions} \cite{ratner2017snorkel}. For intuition, weak supervision is analogous to assigning labels by combining answers from multiple annotators, where each annotator provides a label and a corresponding confidence level. The final label is then a weighted average of the predicted labels. The weights are proportional to (re)calibrated confidence,
where the calibration is used to adjust for miscalibrated annotators. 
In \elicit{}, we use different automated labeling methods as `annotators'. Instead of a single final label, we produce a list ranked by average calibrated confidence, for the human to validate (see \Cref{apx_sec:ranking} for details). 
Beyond its flexibility, a key advantage of weak supervision for our use-cases is that it can enable high overall accuracy even if the labeling functions are not individually performant.

\subsection{Core features and user interface}

\noindent \looseness=-1
\textbf{Leveraging large language models.}
In \elicit{}, the user defines a set of questions (e.g., `Was the victim vulnerable?'), and an ensemble of \textbf{labeling functions} (ranging from keyword lookup to neural nets; see Setup in \Cref{fig:overview}). 
Each labeling function is tasked with 1) identifying a part of the text relevant to the question; and 2) suggesting an answer label for the user to validate (e.g., `Victim was vulnerable.'). 
The most successful labeling functions we tested utilize large language models \citep{devlin2018bert,brown2020gpt3,hoffmann2022chinchilla} to achieve one or both of the above tasks. 
Large language models
can achieve impressive results on passage retrieval and information extraction tasks \citep{agrawal2022large, huang2022finbert}, although not without limitations and biases \citep{bender2021dangers}.    

\vspace{0.5em}
\noindent \looseness=-1
\textbf{Providing explanations.}
To enable human verification, 
label predictions must be accompanied by \emph{explanations}.
For the tasks we consider (\Cref{sec:use_cases}), explanations are equivalent to a relevant snippet of the original text. 
If the provided snippet is insufficient, the user can open a pop-up window containing a larger section of the text. Presenting only relevant snippets is not only faster than manual reading, but also shelters the user from sensitive, graphic, and otherwise problematic content. This reduces both harm to the user, and their mental fatigue.

\begin{figure*}[t]
    \centering
    \includegraphics[width=0.7\textwidth]{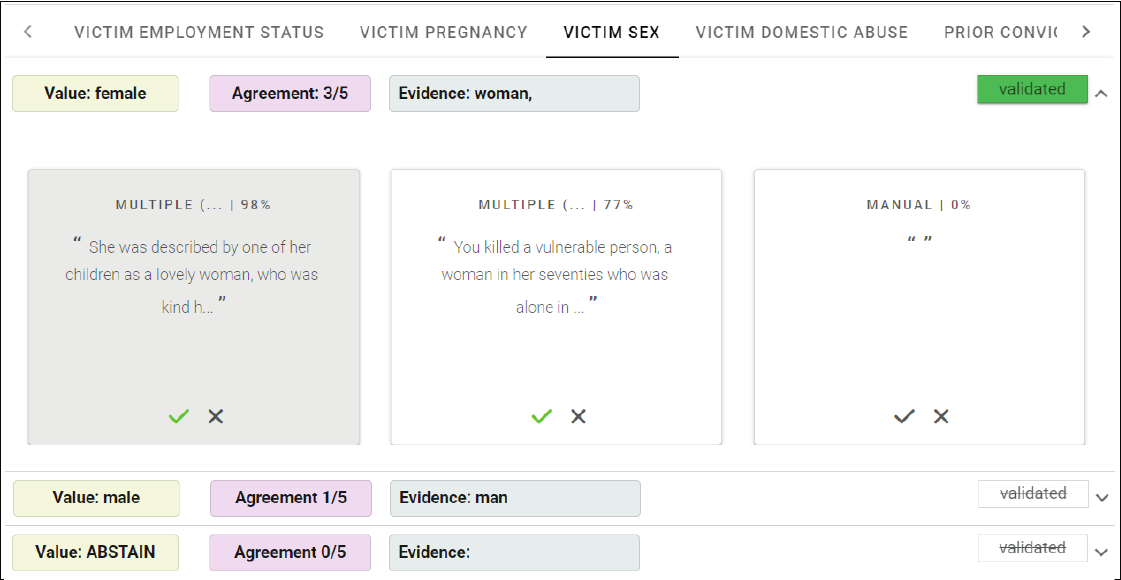}
    \caption{\textbf{User interface screenshot.} 
    In this example, the user is asked to validate that the victim sex. 
    The user is presented with two snippets of the text as explanations. 
    The reported level of agreement corresponds to the maximal LF agreement for a single explanation for each label. 
    The full explanations are presented to the user in a pop-up window, following a mouse click on the box.}
    \label{fig:screenshot}
\end{figure*}

\vspace{0.5em}
\noindent \looseness=-1
\textbf{User interface.}
The user validates candidate labels---e.g., `Victim is female.'---within the user interface (\Cref{fig:screenshot}). 
Too many candidates may overwhelm the user. 
To streamline use and reduce fatigue, we \emph{merge predictions} of the same label if their explanations (snippets) significantly overlap. 
In the \Cref{fig:screenshot} example, the user is asked to validate the victim sex.
Rows correspond to suggested label values. 
The extended snippets for `female' are unrolled. 
You can see that the explanation with the highest confidence was highlighted by 3-out-of-5 labeling functions. 

\looseness=-1
Note that each label value (e.g., male, female) can be supported by multiple explanations (e.g., `She was described \ldots'). 
These are ordered in the UI by a \emph{ranking} function, with the highest confidence explanations placed on the left.
Our ranking model is similar to \citep{ratner2019training}, except we compute scores on a per-explanation rather than per-document level (see \Cref{fig:differences} and \Cref{app_sec:ratnel} for details). 
See supplementary materials for a video demonstration of \elicit{}'s user interface.

\vspace{0.5em}
\noindent \looseness=-1
\textbf{Top-$K$ validation.}
To improve recall, labeling functions can nominate up to \emph{$K$ candidate} tuples---\texttt{(label value, confidence score, explanation)}---for each label, instead of one per document (see Annotation in \Cref{fig:overview}). 
We refer to this as \emph{top-$K$ validation}.
Increasing $K$ improves recall, but can increase burden on the user.




\subsection{Advanced features}
\label{sec:adv}

\noindent \looseness=-1
\textbf{Continual adaptation of candidate ranking.}
As the user validates the extracted information, a new tabular dataset is created. We can use this newly created data to calibrate the confidence output by the labeling functions, and continually \emph{adapt the ranking function to user needs} as in \citep{weasul} (see \Cref{app_sec:ratnel} for details). This will improve the chance that the correct answer will be presented to the user first, making the validation even faster.

\vspace{0.5em}
\noindent \looseness=-1
\textbf{Fine-tuning labeling functions.}
Beyond adapting the ranking, the data provided by user feedback can be used to \emph{continually improve (fine-tune) the labeling functions}.
This can be especially useful for improving recall among the automatically generated candidates.
If a labeling function is updated and a new candidate is found for an already labeled document, \elicit{}'s user interface alerts the user to this fact.

\vspace{0.5em}
\noindent \looseness=-1
\textbf{Deferral.}
\elicit{} can be combined with \emph{deferral} (\Cref{sec:deferral}), where only candidates with low assigned confidence are validated by the user.
This allows further trade-off between time and accuracy, and will work well only if the confidence scores are well calibrated. 
In \Cref{sec:evaluation}, we assume the automated tool comes with its own confidence score, which is used to decide whether to defer to the user via \elicit{}.
We note that the automated extraction algorithm need \emph{not} be one of the \elicit{} labeling functions. 
This allows {specialization}, i.e., fine-tuning one system for automated extraction (e.g., SNORKEL, \citealp{ratner2017snorkel}), and \elicit{} for assisting the user (calibrated scores \& explanations, high recall, etc).

\section{Evaluation}
\label{sec:evaluation}

\looseness=-1
We describe the datasets (\Cref{sec:datasets}),
information to be extracted (\Cref{sec:ie_target}), 
and \elicit{}'s labeling functions (\Cref{sec:elicit_lfs}).
Evaluation of \elicit{}'s core features is presented in \Cref{sec:results}, and of its advanced features in \Cref{sec:results_adv}.
ELICIT and the evaluation code will both be made available on GitHub.

\subsection{Setup}
\label{sec:experimental_setup}

\subsubsection{Data sources}
\label{sec:datasets}

For Task~1 (\Cref{sec:task_crime}), 
we use two self-compiled datasets.
Both contain information on criminal trials for the offence of murder in the United Kingdom. 
For Task~2 (\Cref{sec:task_grooming}), we utilise an existing dataset.

\vspace{0.5em}
\noindent \looseness=-1 \textbf{Crown Court Sentencing Remarks.}
\emph{Sentencing remarks} are a transcript of 
judge's remarks delivered when announcing a sentence.
They usually include a summary of the offence, and sentence justification (incl.\ mitigating and aggravating circumstances). 
The United Kingdom Judiciary publishes sentencing remarks for cases within the realm of public interest \citep{judiciary}. 
There are 343 published sentencing remarks, covering a range of offences, including murder, manslaughter, and misconduct in a public office. 
The vast majority are murder cases; we filter out other offence in the dataset.
To facilitate evaluation,
we manually extracted 18 variables from 20 sentencing remarks, ranging from 500 to 14,000 words in length.
In \Cref{sec:results},
we only report results for variables which occurred in the 20 remarks 5 or more times.

\vspace{0.5em}
\noindent \looseness=-1 \textbf{News articles.}
\emph{Law Pages} is a legal resource website which allows the general public to search for sentencing information, filtering for certain types of offences \citep{thelawpages}. 
The website also links to news articles relating to each case.
Using the filtering tool, we constructed a dataset of sentencing metadata for murder cases tried in the Crown Court, linked to unstructured text from corresponding news articles. 
We only include articles from the five most common outlets (see \Cref{si_fig:press_distribution} in the appendix). 
For validation,
we manually labeled 20 cases
using the relevant subset of variables from the sentencing remarks (10 out of the original 18 variables; see \Cref{si:variable_table} in the appendix).

\vspace{0.5em}
\noindent \looseness=-1 \textbf{Perverted Justice.} 
For Task~2 (\Cref{sec:task_grooming}), 
we use data from the Perverted Justice website \citep{pjwebsite},
which contains real-world chat-based conversations between adults later convicted of grooming offences, and adult decoys posing as children.
We use the 24 annotated chats with 10 variables mentioned in \Cref{sec:task_grooming}, which originate from \citep{Cook2022}.
5 of the 24 chats are used for training and fine-tuning, leaving 19 for evaluation in \Cref{sec:results}.

\subsubsection{Information targeted for extraction}
\label{sec:ie_target}

For each task (\Cref{sec:use_cases}), 
we extract a different set of variables.

\vspace{0.5em}
\noindent \looseness=-1
\textbf{Task~1} (\Cref{sec:task_crime}): 
Information we extract about the \emph{victim}:
race, sex, religion, sexual-orientation, employment status, pregnancy, disability,  
and whether the victim was considered vulnerable,
or suffered physical, mental or domestic abuse at the hands of the defendant.  
Information we extract about the \emph{defendant}:{\footnote{The defendant's demographics are not extracted as these are usually collected as admin data.}
prior convictions, relationship to the victim, sexual or racial motivation, offence premeditation, remorse, and whether age was a mitigating factor.}  

\vspace{0.5em}
\noindent \looseness=-1
\textbf{Task~2} (\Cref{sec:task_grooming}): 
For each conversation, we extract the ten indicator variables described in \Cref{sec:task_grooming}.

\subsubsection{\elicit{}'s labeling functions}
\label{sec:elicit_lfs}

\looseness=-1
We employ distinct LFs for each of our use cases (\Cref{sec:use_cases}).
The \emph{same} LFs are used in the SNORKEL baseline \citep{ratner2019training}, which employs an automated algorithm instead of a human for label validation. 

\vspace{0.5em}
\noindent \looseness=-1
\textbf{Task~1} (\Cref{sec:task_crime}): 
A short description of the five labeling functions we use is below; see \Cref{si:schema_examples} for details. 
\begin{enumerate}[label=LF\arabic*]
    \item \label{lf:qa_zero} \looseness=-1
    \textbf{Transformer Q\&A $\rightarrow$ Zero-shot Sequence Classifier.}
    We use RoBERTa fine-tuned for question-answering on the Squad2 dataset \citep{rajpurkar2018know, liu2019roberta}. 
    The question-answering capability is then used to retrieve relevant text excerpts by associating each variable of interest (e.g., victim sex) with one or more questions (e.g., `What sex was the victim?').
    The model outputs excerpts with associated probabilities $P_\text{QA} (\text{excerpt} \mid \text{question})$.
    We then use a RoBERTa Natural Language Inference (NLI) model \citep{yin2019benchmarking} to assign labels $P_\text{NLI} (\text{label} \mid \text{excerpt})$.
    The score we use to select the top-$K$ candidates is the product
    $P_\text{NLI} (\text{label} \mid \text{excerpt}) \cdot P_\text{QA} (\text{excerpt} \mid \text{question})$.

    \item \label{lf:qa_similarity} \looseness=-1
    \textbf{Transformer Q\&A $\rightarrow$ Similarity.}
    The relevant sections are extracted as in the first step above.
    In the second step, we replace the NLI model by an alternative scoring rule: the cosine similarity $S_{\cos}$ between RoBERTa embeddings of the excerpts and the labels.
    The top-$K$ then maximize $S_{\cos}( \text{excerpt}, \text{label}) \cdot P_{\text{QA}}( \text{excerpt} \mid \text{question} )$. 

    \item \label{lf:similarity} \looseness=-1
    \textbf{Sentence-level similarity.}
    We use a transformer to embed \emph{every} sentence and label.
    Cosine distance is then again used to identify semantic similarities.
    Sentences with similarity above a user-defined threshold are extracted.
    The top-$K$ sentences with similarity above a user-defined threshold are taken as candidates.
    
    \item \label{lf:keyword}
    \textbf{Keyword Search.}
    Each label is associated with a set of user-specified keywords.
    For example, for `victim sex', the label value `male' could be associated with keywords like `man', `male', `Mr.', and so on. In this case, top-$K$ does not apply due to no clear way of assigning scores to different candidates.
    
    \item \label{lf:vqa_zero}
    \textbf{Visual Q\&A $\rightarrow$ Zero-shot Sequence Classifier.}
    Same as \ref{lf:qa_zero},
    except with RoBERTa replaced by LayoutLM \citep{layoutlm},
    a visual question-answering model fine-tuned on the Squad2 and DocVQA datasets \citep{rajpurkar2018know, docvqa}.
    LayoutLM operates directly on PDFs, making use of layout and visual information. The top-$K$ candidates are obtained as in \ref{lf:qa_zero}.
    
\end{enumerate}

\vspace{0.5em}
\noindent \looseness=-1
\textbf{Task~2} (\Cref{sec:task_grooming}): 
We modify \ref{lf:qa_zero},
\ref{lf:qa_similarity},
and \ref{lf:similarity} from above, 
and add one new LF (referred to as \ref{lf:nli_pretrained} for coherence).
For \textbf{\ref{lf:qa_zero}} and \textbf{\ref{lf:qa_similarity}}, the key difference are the questions for the Q\&A transformer.
Here, we derive these from the original coding dictionary \citep{Cook2022}.
For example, for \emph{rapport building} (one of the ten indicators to be extracted; see \Cref{sec:use_cases,sec:datasets}), 
the questions we use start with 
\textit{`Is the offender \ldots ?'} with the ellipsis substituted by:
(i)~\textit{`giving a compliment'};
(ii)~\textit{`accepting a compliment'};
(iii)~\textit{`building a special bond'};
(iv)~\textit{`being romantic'};
(v)~\textit{`showing interest'};
(vi)~\textit{`talking about personal similarities'}.
The questions used for the other 
indicators can be found in \Cref{si:schema_examples}. 
For \textbf{\ref{lf:similarity}} the only modification is 
setting the categories to the ten indicators of interest (\Cref{sec:task_grooming}).

\looseness=-1
\begin{enumerate}[label=LF6]
\item \label{lf:nli_pretrained}
\textbf{Pre-trained NLI sequence classifier}.
We use a fine-tuned version of RoBERTa-large for NLI obtained from \citep{Cook2022}. 
For all offender's messages,
the classifier assigns ten scores, each specifying how model's confidence that given behaviour of interest (\Cref{sec:task_grooming}) is present.
Predictions with confidence of at least $0.4$ can be flagged for user validation.
If there are more than $K$ such predictions (see \Cref{sec:elicit}), the $K$ highest confidence messages per each of the ten labels are returned.
If no messages exceeded the threshold, `no evidence' is returned. 
\end{enumerate}

\subsection{Results: core features}
\label{sec:results}

\begin{figure}[tb]
    \centering
    \includegraphics[width=0.6\columnwidth]{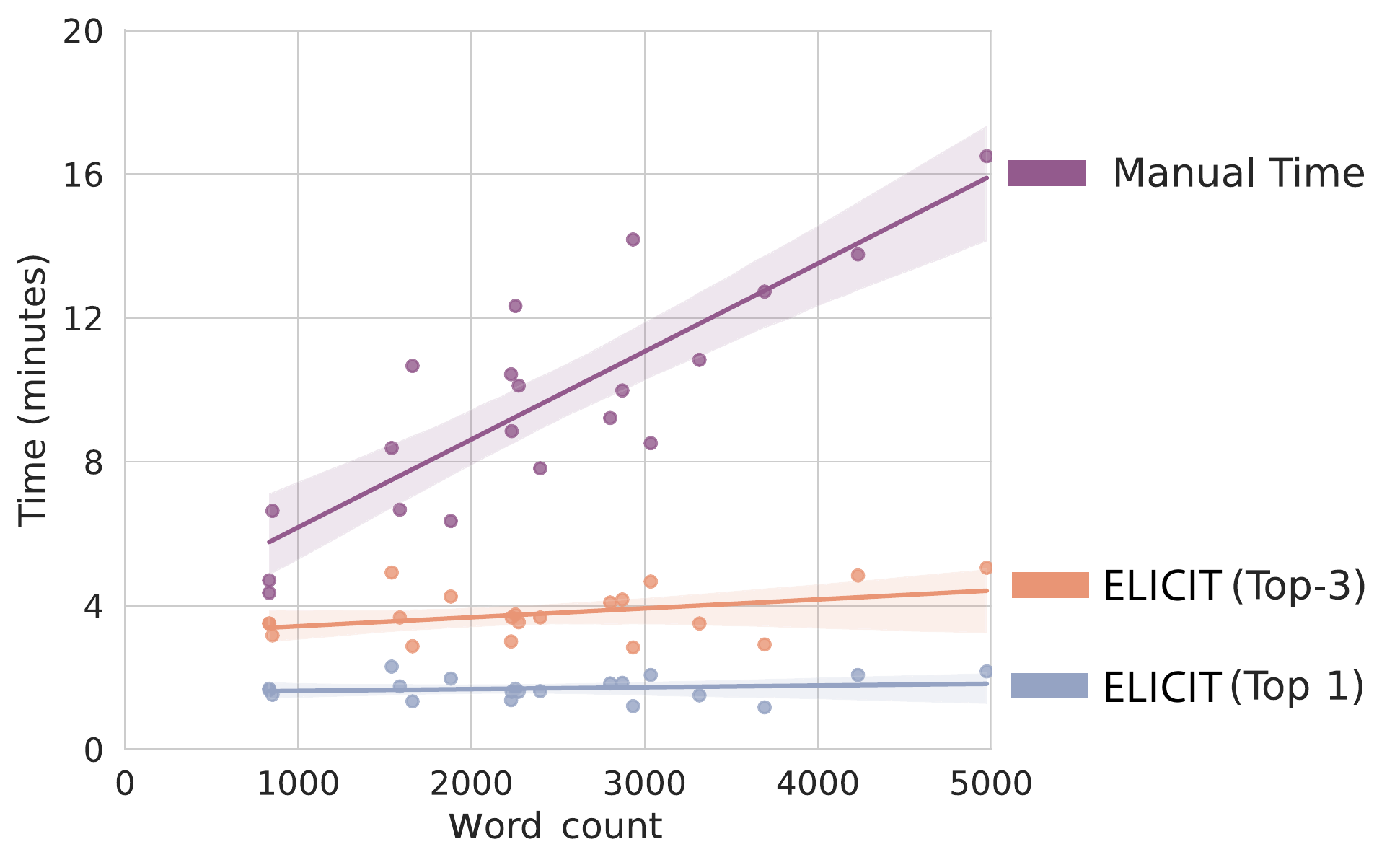}
    \caption{\looseness=-1
    \textbf{Annotation time vs.\ word count on Sentencing remarks.} 
    For top-1 ELICIT (blue), time is constant, whereas manual annotation (purple) scales roughly linearly with word count. Increasing the allowed answers per LF from 1 to 3, doubles the annotation time.}
    \label{fig:sr_time}
\end{figure}

\subsubsection{Time-efficiency}

\looseness=-1
The difference in annotation times between manual and semi-automated extraction depends on length of the documents. 
\Cref{fig:sr_time} shows the annotation times for top-1 and top-3 ELICIT vs.\ manual annotation, on the Sentencing remarks dataset (\Cref{sec:datasets}).
Top-1 and top-3 ELICIT respectively took 102 and 227 seconds on average, \emph{independently of the document length}.
Compared to manual annotation, top-1 ELICIT achieved an order-of-magnitude speed-up for documents containing 4000 or more words.
For Task~2, annotators only timed the overall time.
On average, each conversation took 2.4 minutes to annotate, achieving a $\sim$20x time reduction compared to manual annotation. 
Top-1 and top-3 achieved similar timing, as the annotator was more familiar with the system when preforming the top-3 validation. 

\begin{figure*}[tb]
    \centering
    \includegraphics[width=\textwidth]{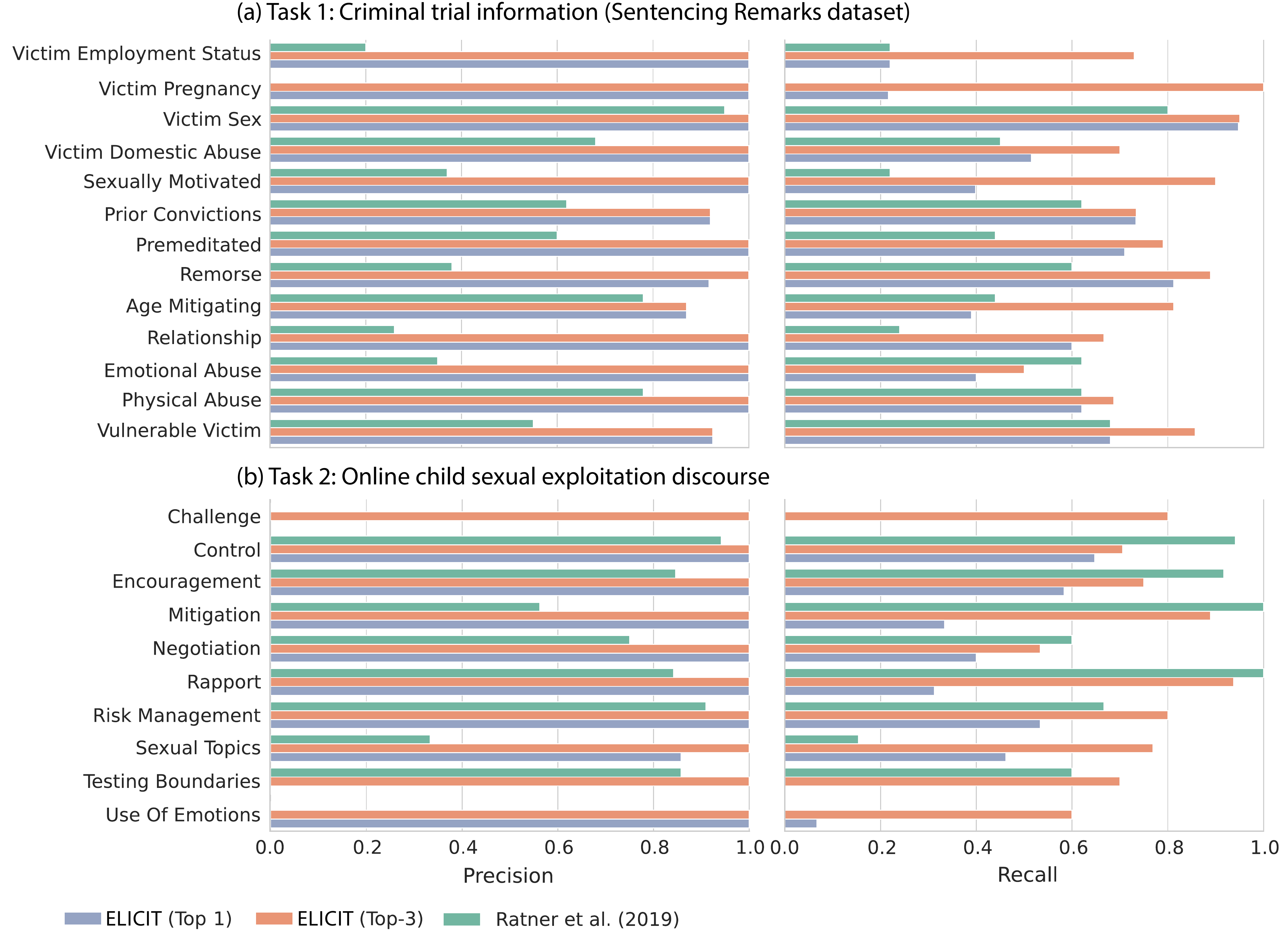}
    \caption{\looseness=-1
    \textbf{Weighted precision and recall performance.} 
    Top-1 (blue) and Top-3 (orange) ELICIT is compared to fully automated extraction from \citep{ratner2019training}. Precision and recall are weighted by per-class support as to not misrepresent the performance due to class imbalance.}
    \label{fig:sr_perf}
\end{figure*}

\subsubsection{Performance}

\looseness=-1

We compare the precision and recall of 
\elicit{} to an automated baseline
called
SNORKEL \citep{ratner2017snorkel} 
(see \Cref{apx_sec:maj_rule} for a majority rule baseline). 
In \Cref{fig:sr_perf} we show the comparison of top-K validation in \elicit{}, using $K = 1, 3$.
As expected, for both tasks, the {precision} of \elicit{} is comparable to manual extraction, irrespective of $K$.\footnote{Top-1 \elicit{} did not retrieve any `Challenge' label candidates, which is reported as zero precision.}

For Task~1,
the user incorrectly validated only five (out of 260) of the instances.
The failures were due to contradictions within the retrieved snippets.
For example, the user selected `not premeditated' after 
seeing the quote
``I cannot find that there was a lack of premeditation such as to amount to a mitigating factor''. 
However, reading the full text, the judge later added that ``Whilst I do not find there was a significant degree of planning or premeditation as an aggravating factor, equally I cannot find that there was a lack of premeditation such as to amount to a mitigating factor.''

\looseness=-1
For full automation,
precision varied among the extracted variables.
It fell below $0.5$ for 4/13 (Task~1), resp.\ 3/10 (Task~2), of the variables.
Precision of both models was better on Task~2,
likely due to addition of LF6.
ELICIT's overall superior precision is evidence that the retrieved text snippets provide the user with sufficient context for label validation.

Increasing the number of candidates from top-1 to top-3 had little impact on precision but it significantly improved in \emph{recall}. 
For Task~1, top-1 performed slightly better than the automated baseline with 0.56 and 0.46 mean recall, while top-3 outperforms both with 0.78 mean recall.
The $K$-related recall improvement is even more pronounced for Task~2,
where the automated baseline outperformed top-1 ELICIT with 0.58 vs.\ 0.33. 
However, top-3 ELICIT increased recall to 0.75, surpassing both.
In ELICIT, the high accuracy of the user implies that low recall only occurs when relevant information is not retrieved by the LFs (modulo user fatigue).
The large recall improvement when moving from top-1 to top-3 thus suggests that the LF confidence is not sufficiently well-calibrated.
Similarly to other weak supervision methods, recall can be further improved by adding and improving the LFs (e.g., by adding keywords or paraphrased questions). 
\Cref{fig:sr_perf} only reports performance only on the Sentencing Remarks dataset. 
We report the precision and recall for the press articles dataset (Task~1) in \Cref{si_fig:lawpages_perf} in the appendix.

\subsection{Results: advanced features}
\label{sec:results_adv}

\begin{figure}[tb]
    \centering
    \includegraphics[width=0.6\columnwidth]{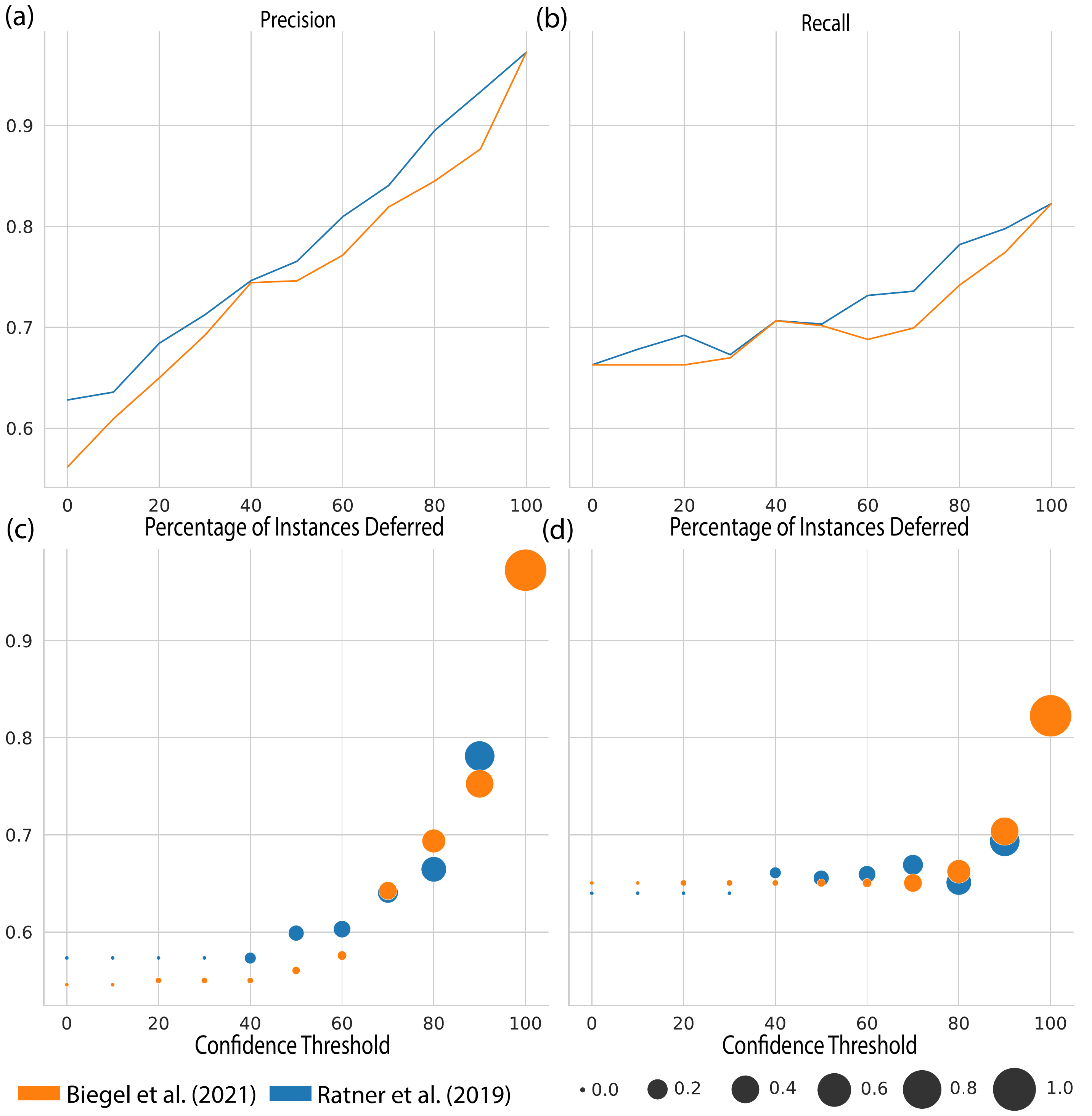}
    \caption{
    \textbf{Performance under different deferral schemes.} (a, b) Precision and recall as a function of percentage of instances deferred to ELICIT.
    The lowest confidence instances are deferred first.
    (c, d) Precision and recall when instances below a given confidence threshold are deferred from to ELICIT. 
    Point sizes are scaled by the number of instances under a given confidence threshold. }
    \label{fig:deffral}
\end{figure}

\subsubsection{Validation vs. deferral} 

\looseness=-1
When time efficiency is critical, it may be worth considering a more efficient solution than top-1 validation. 
We use the Sentencing Remarks data (\Cref{sec:datasets}) to demonstrate how performance changes for two types of deferral:
1)~fixed \emph{budget} deferral, where only a certain percentage of the cases can be deferred to a human; and 
2)~fixed \emph{threshold} deferral, where every prediction below a certain confidence is deferred to a human. \Cref{fig:deffral} shows the trade-off in precision and recall for fixed budget (a \& b) and threshold (c \& d).
The colours distinguish between ranking methods with \citet{ratner2019training} in blue, and \citet{weasul} in orange.
Both achieve similar results in our setting.

\looseness=-1
Precision improves linearly with percentage of deferred cases. 
Recall, however, improves faster when more cases are human validated.
In this setting, a saving of 40\% in human time will result in a loss of $\sim$20\% precision and $\sim$10\% recall, highlighting the value of adding a human-in-the-loop element. 
Deferral may be the preferred option in settings where either
1)~the difference in performance between the automation and human validation is small,
2)~the confidence produced by the automated methods is well-calibrated, or 
3)~reducing time and human-effort is a higher priority. 

\begin{figure}[tb]
    \centering
    \includegraphics[width=0.6\columnwidth]{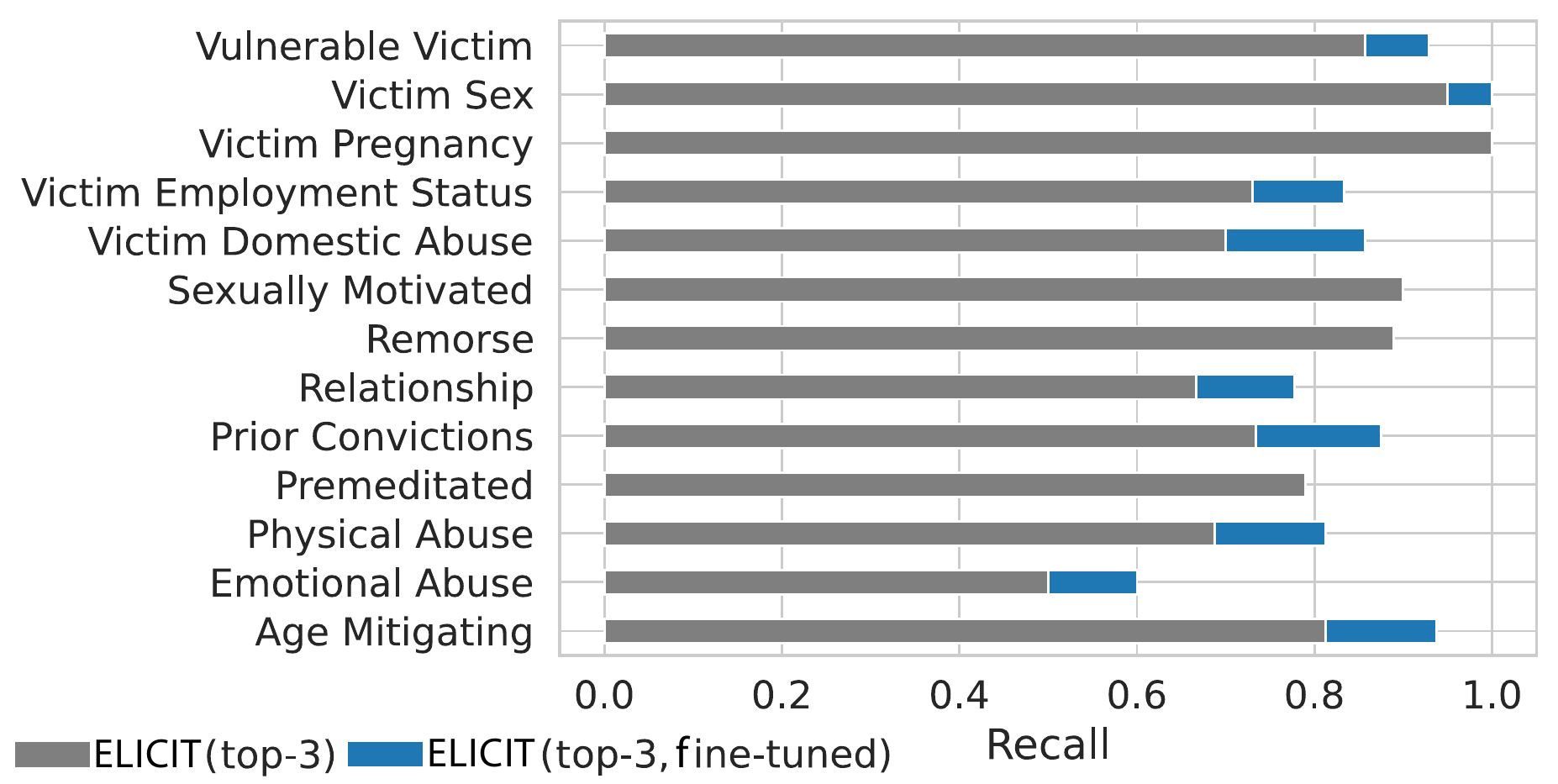}
    \caption{\textbf{The recall delta before and after fine-tuning with ELICIT} on the Sentencing remarks dataset.}
    \label{fig:training}
\end{figure}

\subsubsection{Improving recall via training} 
\label{sect:training_from_feedback}

\looseness=-1
With every human validation, we obtain an additional label beyond the original dataset.
This can be used to improve recall by fine-tuning the labeling functions.
Beyond improved performance on new instances, \elicit{} also alerts the user if fine-tuning resulted in new validation opportunity in already reviewed data.

\Cref{fig:training} shows the improvement in recall when fine-tuning on 20 validated cases.
The average improvement in recall was 0.075, with a maximum improvement of 0.15 (for domestic abuse, 0.7 increased to 0.85). 
These auxiliary uses of the validated data---refining the ranking, and fine-tuning the LFs---may improve performance enough to make deferral or even fully automated extraction feasible in the longer term in certain settings.

\subsubsection{Reflections from the user's perspective}

One of the goals behind developing \elicit{} was to improve the well-being of human annotators. 
Although we did not measure well-being quantitatively, 
we documented reflections from the two authors who
annotated \emph{both} manually and using \elicit{}.
For Task~1, the author who extracted information about murder cases tried in the UK, said: `Reviewing large amounts of court materials in such detail is mentally draining. I found it could limit the quantity which I could label in a single sitting. Using ELICIT limits the exposure to unnecessary material, allowing more labelling to be completed before a break is required.'
The annotator for Task~2, who is an experienced annotator, found the user interface improved their own ability to perform the task, stating that `Manual message-by-message labelling is difficult because it does not really reflect how humans think naturally, while chunking the messages together and asking me ``Is this rapport?'' seems more familiar as an everyday task.'
This annotator stated that while the content was equally unpleasant to read, it was a relief to spend less time with it.

\section{Ethics and social impact} 
\label{sec:ie_users}

\paragraph{Users.} The evaluations presented in this paper were conducted by the authors of the paper and their collaborators. Specifically, for Task~1, manual extraction for 25 cases was done by two authors, one of whom also completed the \elicit{} extraction for 100 cases.
For Task~2, manual extraction from 66 conversations was done by two forensic psychologists as part of their MSc dissertations. 
Validation and \elicit{} extraction was conducted by an author who is also a forensic psychologist. 
Ethics clearance was received, separately, for both tasks, via the participants' departmental ethics board.
Due to the potentially traumatizing nature of the data, no external participants were recruited. 

\paragraph{Data.} All the data used in this study are publicly available. For Task~1, although the original sentencing remarks and news articles contained full names, we removed these when creating the datasets. These could be re-identified by finding the original publication; however, as the data is in the public domain, this is unlikely to result in further harm to the individual's privacy. 
For Task~2, the data is publicly available via the Perverted Justice website \citep{pjwebsite}. 
The website contains conversations between adults and adult decoys pretending to be minors. 
No actual minors were involved in the conversations. 
The website only publishes chats that have led to convictions. 
We note that it is highly unlikely that the offenders consent to publish these conversations. 
While contestable, this data is the only publicly available source in the domain, and has been widely used in the relevant literature.

\paragraph{Potential impact of this research.} \looseness=-1
The goal of this work is to facilitate creation of new tabular datasets, which would enable new avenues of quantitative and mixed-methods research. This work opens new routes to fill critical data gaps, which is much needed to improve our understanding of the criminal justice system \citep{Zilka2022datasets}. 
However, we acknowledge our research can lead to creation of low-quality, misleading, or cherry-picked datasets, if used irresponsibly.
We stress the necessity of accompanying any created dataset by detailed extraction methodology documentation, including how ambiguity is dealt with, true/false positive/negative rates from sample testing, and disclosure of any other known biases and errors.

\section{Conclusion}
\label{sec:conc}

\looseness=-1
We present a framework for human-validated information extraction from text.
To ensure real-world grounding, we centred our investigation around two use cases from the criminal justice domain.
Based on their commonalities, we identified several key functionality requirements (accuracy, time-efficiency, flexibility, reproducibility, and reducing user's exposure to harmful content).
We reviewed suitability of different collaborative settings and tools with respect to the identified requirements.
Since none satisfied all our requirements, we developed \elicit{} which we release as open-source.

\looseness=-1
\elicit{} is a flexible tool useful for a variety of information extraction tasks.
Its design is inspired by weak supervision approaches: we use a set of algorithmic annotators (labeling functions) to identify relevant pieces of text, 
which are then validated by the user.
Compared to manual annotation, \elicit{} can attain comparable accuracy at fraction of time.
This is achieved by leveraging the complementary strengths of humans and machines in our use cases: speed and accuracy.

\looseness=-1
We evaluate \elicit{} on three extraction tasks based on our criminal justice use cases. In all cases, we achieve accuracy close to manual annotation with \emph{orders of magnitude lower time investment}.
\elicit{} significantly outperforms the \emph{automated} extraction on both precision and recall.
We demonstrate that recall can be further improved by using the already validated data for fine-tuning. 
We further quantify the trade-off between human effort and performance within a deferral setup. 
Finally, based on our own experience extracting information manually and using \elicit{}, we found that using \elicit{} required less emotional strain compared to manual annotation.

\looseness=-1
Our framework can be particularly effective for extraction of factual information from very long documents, when high precision is an essential requirement. 
Beyond helping human annotators do their work more effectively, we believe the continual learning component of our system (\Cref{sect:training_from_feedback}) is a promising direction for improving machine performance via human feedback.
Learning from human feedback is a topic of growing importance in machine learning \citep[e.g.,][]{jeon2020reward,griffith2013policy,christiano2017deep}, where fine-tuning language models on human feedback recently lead to significant gains \citep{ouyang2022training,bai2022training,chung2022scaling,thoppilan2022lamda}.
These approaches largely rely on human annotators hired specifically for the purpose of ranking model outputs based on their quality.
In contrast, \elicit{} uses the interactions with users themselves to become more performant, demonstrating a complementary venue to effectively collecting and learning from human feedback.
We hope \elicit{}'s value-led design---combined with engineering choices informed by task-specific requirements---inspires further research in this direction.


\bibliography{Elicit_Arxiv}
\bibliographystyle{ACM-Reference-Format}

\newpage

\appendix
\onecolumn

\section{Performance}
\renewcommand{\thefigure}{A\arabic{figure}}
\setcounter{figure}{0}

\begin{figure*}[h]
    \centering
    \includegraphics[width=\textwidth]{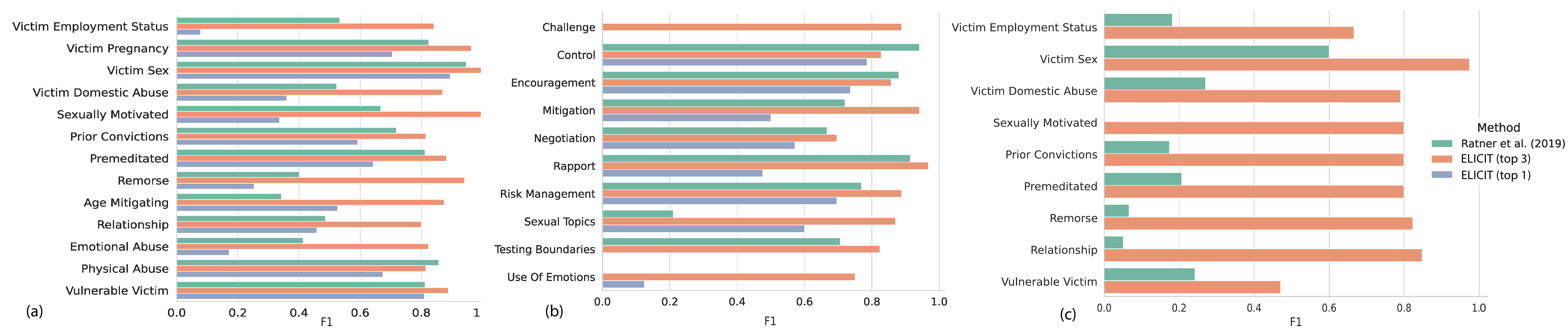}
    \caption{F1 scores all tasks: (a) sentencing remarks, (b) perverted justice, and (c) press articles. 
    Top-1 (blue) and Top-3 (orange) ELICIT is compared to fully automated extraction from \citep{ratner2019training}.}
    \label{fig:sr_perf_f1}
\end{figure*}

\section{Press Articles Dataset}
\renewcommand{\thefigure}{B\arabic{figure}}
\renewcommand{\thetable}{B\arabic{table}}
\setcounter{figure}{0}
\setcounter{table}{0}

\begin{figure}[h]
    \centering
    \includegraphics[width=0.5\textwidth]{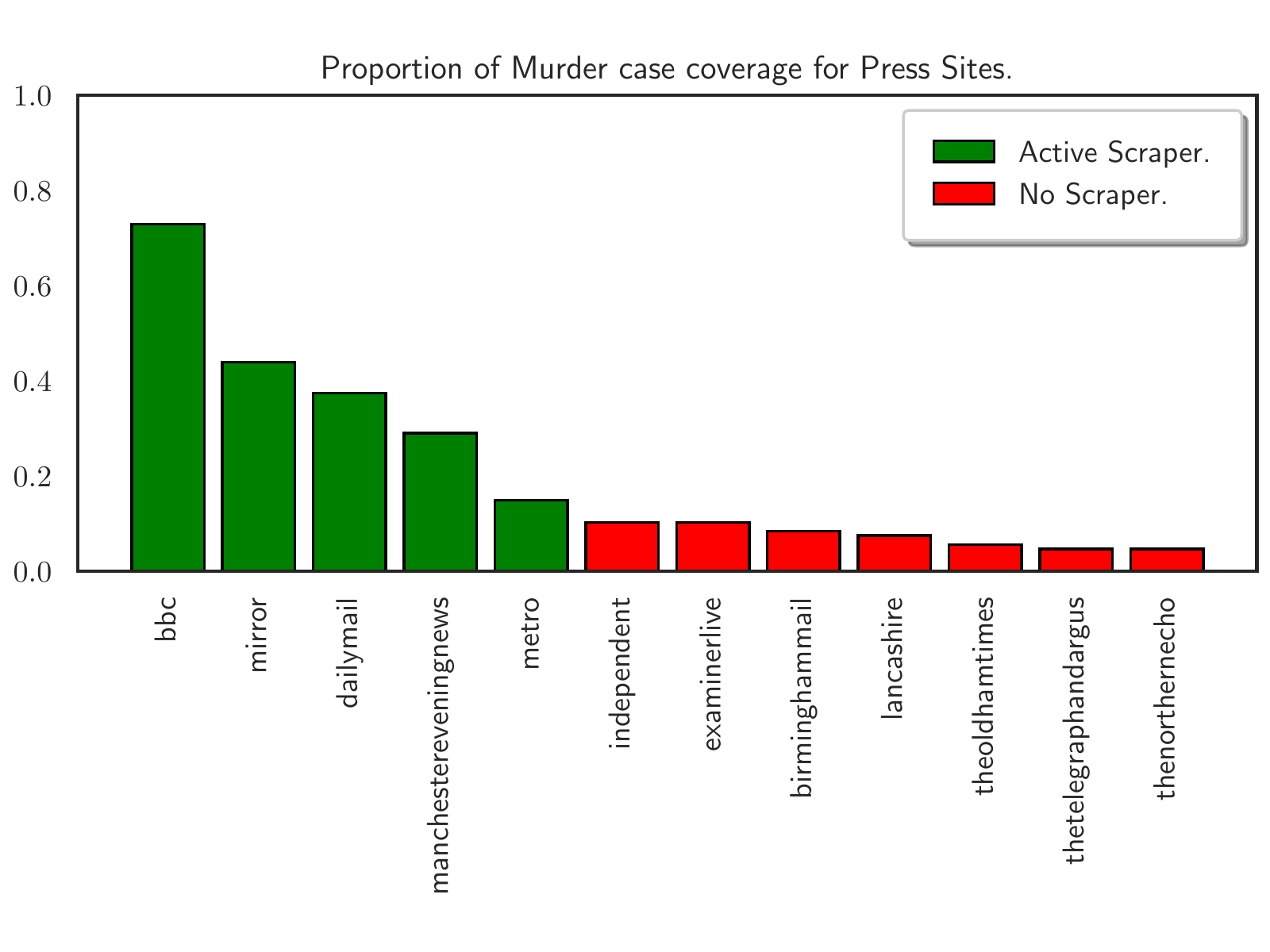}
    \caption{Distribution of press site coverage for murder cases from the Law Page website. Colour indicates whether we use articles from the corresponding sites: (green) use, (red) don't use. The five highest frequency sites are used.}
    \label{si_fig:press_distribution}
\end{figure}

\begin{table}[h]
\centering
\resizebox{0.5\textwidth}{!}{%
\begin{tabular}{lcc}
\hline
Variable                  & Sentencing Remarks & News Articles             \\ \hline
Victim Sex                & \cmark     & \cmark     \\
Victim Domestic Abuse     & \cmark     & \cmark     \\
Victim Vulnerable         & \cmark     & \cmark     \\
Victim Pregnancy          & \cmark     & \xmark     \\
Victim Employment Status  & \cmark     & \cmark     \\ 
Physical Abuse            & \cmark     & \xmark     \\
Mental Abuse              & \cmark     & \xmark     \\
Remorse                   & \cmark     & \cmark     \\
Prior Convictions         & \cmark     & \cmark     \\
Sexually Motivated        & \cmark     & \cmark     \\
Age Mitigating            & \cmark     & \xmark     \\
Premeditation            & \cmark     & \cmark     \\
Relationship              & \cmark & \cmark \\ \hline
\end{tabular}}
\caption{The list of variables, with check marks and crosses indicating whether the variable was extracted for the corresponding dataset. Variables consistently abstained while labeling news articles were removed from the news article annotation process. }
\label{si:variable_table}
\end{table}

\begin{figure*}[h]
    \centering
    \includegraphics[width=0.9\textwidth]{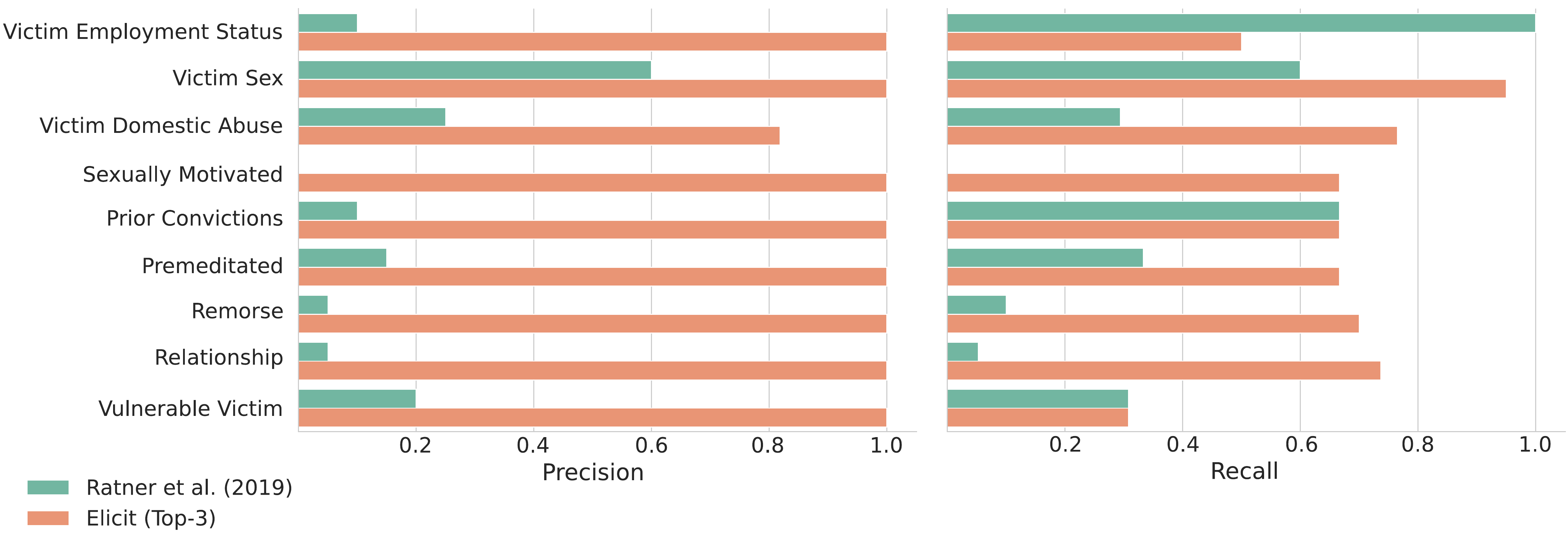}
    \caption{Weighted precision and recall performance on the lawpages dataset. 
    ELICIT (orange) Top-3 is compared to fully automated extraction (green) from \citep{ratner2019training}. Precision and recall are weighted by per-class support as to not misrepresent the performance due to class imbalance.}
    \label{si_fig:lawpages_perf}
\end{figure*}

\begin{figure*}[bh]
    \centering
    \includegraphics[width=0.45\textwidth]{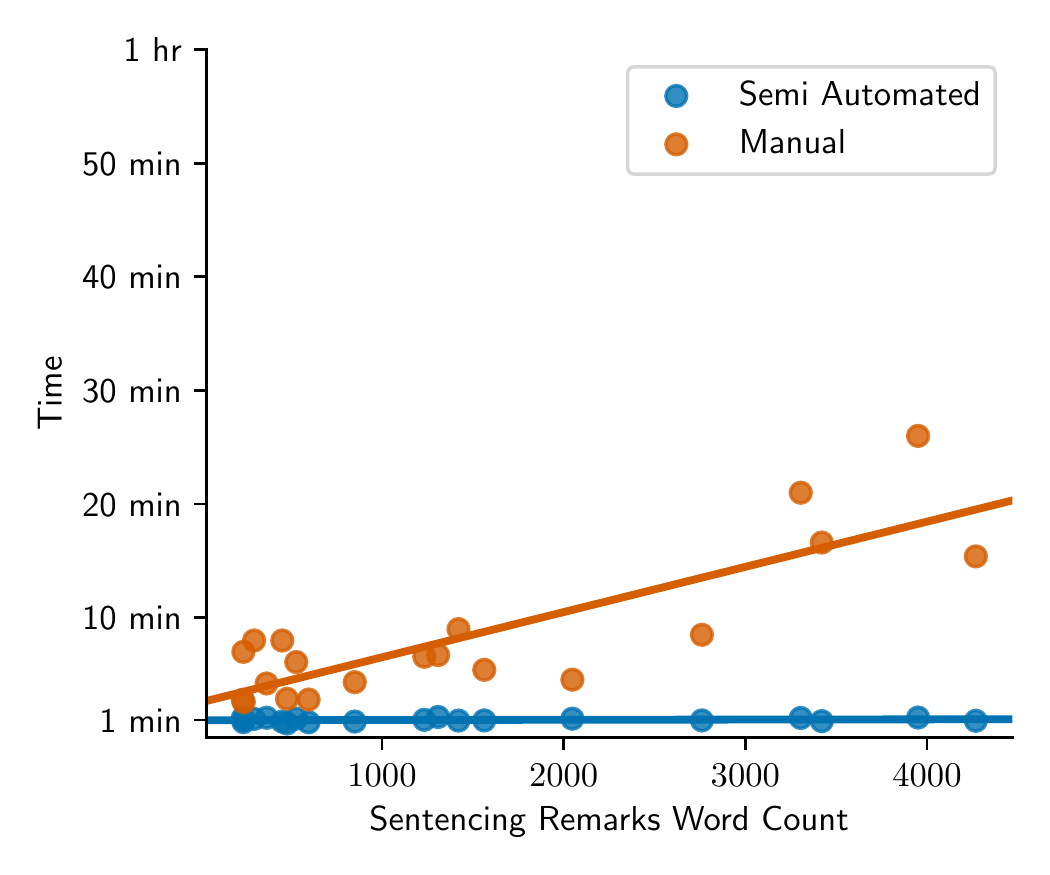}
    \caption{Plot of annotation time against law pages word count. For Elicit semi-automated annotation (blue), time is constant, whereas manual annotation (orange) scales roughly linearly with word count.}
    \label{si_fig:lawpages_time}
\end{figure*}
\FloatBarrier

\newpage

\section{Majority Rules}
\label{apx_sec:maj_rule}
\renewcommand{\thefigure}{C\arabic{figure}}
\renewcommand{\thetable}{C\arabic{table}}
\setcounter{figure}{0}
\setcounter{table}{0}

\begin{figure*}[h]
    \centering
    \includegraphics[width=\textwidth]{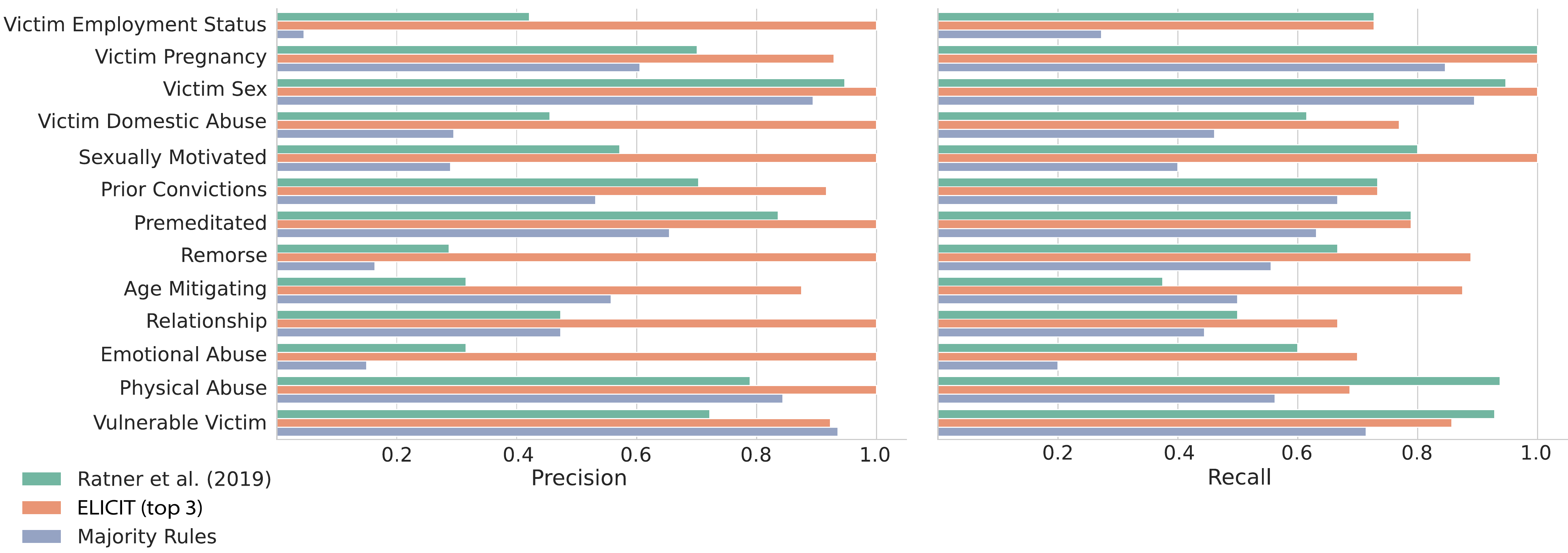}
    \caption{Weighted precision and recall performance for sentencing remarks. 
    Majority Rules (blue) and Top-3 (orange) ELICIT is compared to fully automated extraction from \citep{ratner2019training}. Precision and recall are weighted by per-class support as to not misrepresent the performance due to class imbalance.}
\end{figure*}

\FloatBarrier

\section{Schema Examples}

\label{si:schema_examples}

In the following, we provide a small example of the schemas used for the variables \textit{victim sex} and \textit{prior convictions}. The four labeling functions used require three schemas: category, question, and keyword schema. We represent these schemas here as lists; in reality, these are defined in YAML files.

\subsubsection*{Sentencing Remarks: Category Schema}
\begin{itemize}
    \item Victim sex
    \begin{itemize}
    \item Male
    \item Female
    \end{itemize}
    \item Prior Convictions
    \begin{itemize}
    \item Prior Convictions
    \item No Prior Convictions
    \end{itemize}
\end{itemize}

\subsubsection*{Sentencing Remarks: Question Schema}
\begin{itemize}
    \item Victim sex
    \begin{itemize}
    \item What sex was the victim?
    \item Was the victim male?
    \item Was the victim female?
    \end{itemize}
    \item Prior Convictions
    \begin{itemize}
    \item Prior convictions?
    \item Did the defendant have prior convictions?
    \item Previous crimes?
    \end{itemize}
\end{itemize}

\subsubsection*{Sentencing Remarks: Keyword Schema}
\begin{itemize}
    \item Victim sex
    \begin{itemize}
      \item male:
      \begin{itemize}
          \item male
          \item man
          \item boy
      \end{itemize}
      \item female:
      \begin{itemize}
          \item female
          \item woman
          \item girl
      \end{itemize}
    \end{itemize}
        \item Prior Convictions
    \begin{itemize}
      \item Prior Convictions:
      \begin{itemize}
          \item prior convictions
          \item previous convictions
          \item criminal record
      \end{itemize}
      \item No Prior Convictions:
      \begin{itemize}
          \item No prior convictions
          \item Previous good character
      \end{itemize}
    \end{itemize}
\end{itemize}

\subsubsection*{Perverted Justice: Category Schema}
\begin{itemize}
    \item Rapport
    \begin{itemize}
    \item Rapport
    \item No Rapport
    \end{itemize}
    \item Control
    \begin{itemize}
    \item Control
    \item No Control
    \end{itemize}
    \item Negotiation
    \begin{itemize}
    \item Negotiation
    \item No Negotiation
    \end{itemize}
    \item Challenge
    \begin{itemize}
    \item Challenge
    \item No Challenge
    \end{itemize}
    \item Use of emotions
    \begin{itemize}
    \item Use of emotions
    \item No use of emotions
    \end{itemize}
    \item Mitigation
    \begin{itemize}
    \item Mitigation
    \item No mitigation
    \end{itemize}
    \item Encouragement
    \begin{itemize}
    \item Encouragement
    \item No encouragement
    \end{itemize}
    \item Risk Management
    \begin{itemize}
    \item Risk Management
    \item No Risk Management
    \end{itemize}
    \item Sexual Topics
    \begin{itemize}
    \item Sexual Topics
    \item No Sexual Topics
    \end{itemize}
    \item Testing Boundaries
    \begin{itemize}
    \item Testing Boundaries
    \item No Testing Boundaries
    \end{itemize}
\end{itemize}

\subsubsection*{Perverted Justice: Question Schema}
\begin{itemize}
    \item Rapport
    \begin{itemize}
    \item is the offender giving a compliment?
    \item is the offender accepting a complement?
    \item is the offender building a special bond?
    \item is the offender being romantic?
    \item is the offender showing interest?
    \item is the offender talking about personality?
    \item is the offender talking about personal similarities?
    \end{itemize}
    \item Control
    \begin{itemize}
    \item is the offender being persistent?
    \item is the offender talking about consent?
    \item is the offender trying to please the victim?
    \item is the offender complying with requests?
    \item is the offender jealous?
    \item is the offender being compliant?
    \item is the offender being assertive?
    \item is the offender asking a rhetorical question?
    \item is the offender being patronising?
    \item is the offender asking for permission?
    \item is the offender checking for engagement?
    \end{itemize}
    \item Negotiation
    \begin{itemize}
    \item is the offender offering incentives?
    \item is the offender making plans to meet?
    \item is the offender persuading the victim?
    \item is the offender defensive?
    \item is the offender talking about alcohol?
    \item is the offender talking about drugs?
    \item is the offender arranging plans?
    \end{itemize}
    \item Challenge
    \begin{itemize}
    \item is the offender mocking the victim?
    \item is the offender insulting the victim?
    \item is the offender confronting the victim?
    \item is the offender rejecting the victim?
    \item does the victim trust the offender?
    \end{itemize}
    \item Use of emotions 
    \begin{itemize}
    \item is the offender showing concern?
    \item is the offender looking for validation?
    \item is the offender shocked?
    \item is the offender angry?
    \item is the offender sad?
    \item is the offender confused?
    \item is the offender embarrassed?
    \item is the offender happy?
    \item does the offender reassure the victim?
    \item does the offender ask for reassurance?
    \end{itemize}
    \item Mitigation
    \begin{itemize}
    \item does the offender implicate themselves in a crime?
    \item does the offender have a sexual preference for children?
    \end{itemize}
    \item Encouragement
    \begin{itemize}
    \item does the offender express willingness to engage?
    \item does the offender encourage the victim?
    \item does the offender comply with the victim?
    \item does the offender flirt with the victim?
    \item does the offender request a picture of the victim?
    \end{itemize}
    \item Risk management
    \begin{itemize}
    \item does the offender ask if the victim is real?
    \item does the offender ask if the victim is a cop?
    \item does the offender ask about the victim's mom?
    \item does the offender ask about the victim's dad?
    \item does the offender ask about the victim's family?
    \item  does the offender talk about the dangers on the internet?
    \item does the offender ask about meeting the victim?
    \end{itemize}
    \item Sexual Topics
    \begin{itemize}
    \item is the offender talking about sexual topics?
    \item is the offender talking about fantasies?
    \item is the offender talking about sexual preferences?
    \item is the offender talking about pornography?
    \item is the offender talking about sexual acts?
    \item is the offender talking about relationships?
    \item is the offender talking about age differences?
    \end{itemize}
    \item Testing Boundaries
    \begin{itemize}
    \item does the offender set boundaries?
    \item does the offender check the victim's willingness to engage?
    \item does the offender talk about sex?
    \item does the offender talk about relationships?
    \item does the offender talk about sharing pictures?
    \item does the offender talk about meeting offline?
    \item does the offender talk about fantasies?
    \item does the offender talk about sharing pictures?
    \item is the offender being secretive?
    \item is the offender bored?
    \end{itemize}
\end{itemize}

\newpage

\section{Ranking}
\label{apx_sec:ranking}

\renewcommand{\thefigure}{E\arabic{figure}}
\renewcommand{\thetable}{E\arabic{table}}
\setcounter{figure}{0}
\setcounter{table}{0}

\subsection{Differences between \elicit{} and Ratner et al.}

\begin{figure}[ht]
    \centering
    \includegraphics[width=0.8\textwidth]{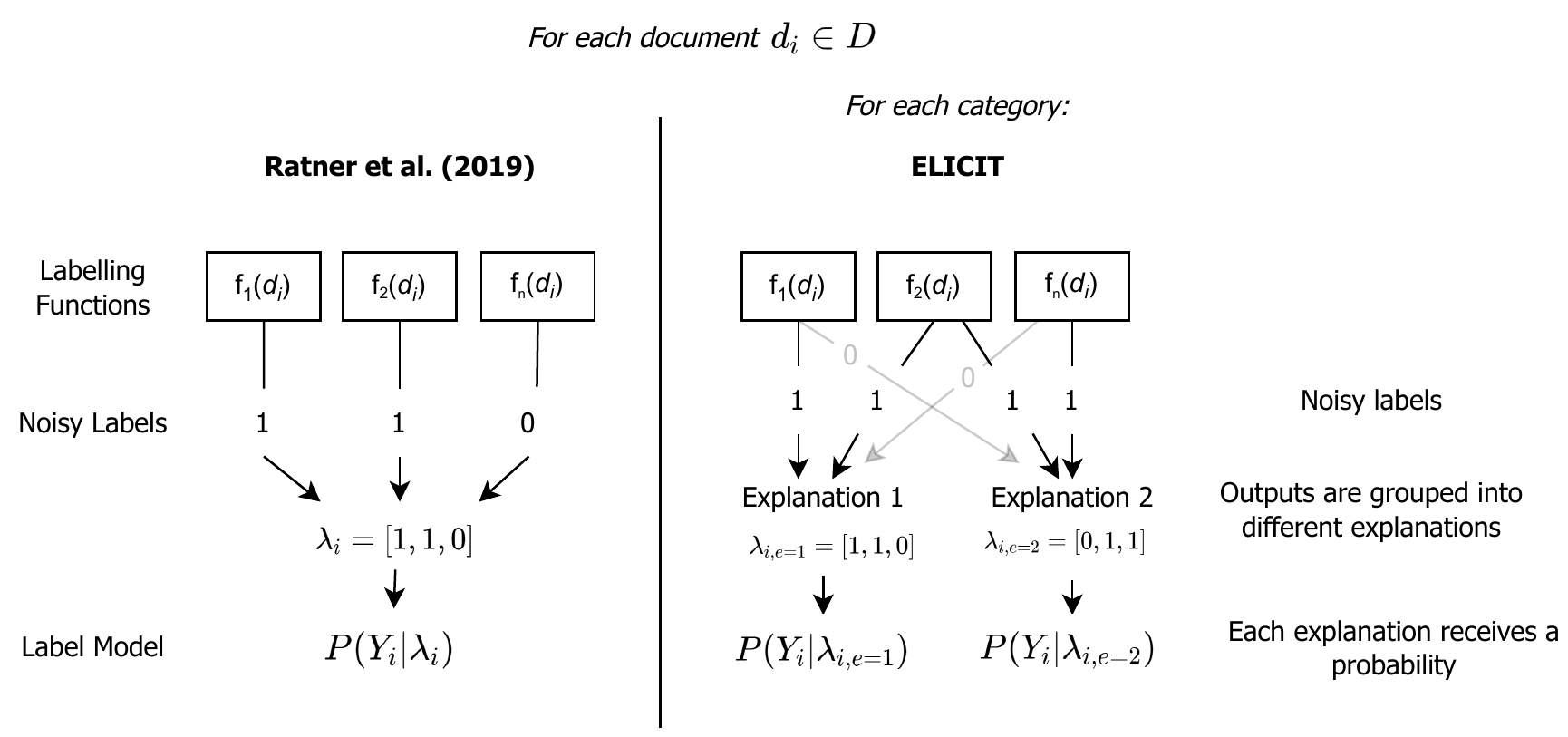}
    \caption{An overview of differences between our method and \citep{ratner2019training}.
    In \citep{ratner2019training}, the labeling functions predict one label value per document, which is then encoded in a vector $\lambda_i$ (eventually a matrix $\lambda$ over all documents). This $\lambda$ is used to learn a generative model $p(Y_i|\lambda_i)$. 
    In our case, the labeling functions can produce multiple answers, each with an associated explanation. 
    Common explanations are grouped into a $\lambda_i$ vector, and corresponding probability $P(Y_i|\lambda_i)$ is produced on a per-explanation, rather than per-document, basis.}
    \label{fig:differences}
\end{figure}

\subsection{Ratnel et al.}
\label{app_sec:ratnel}

\looseness=-1
We use the weak supervision method presented in \citep{ratner2019training} in order to get a confidence for each explanation. 
We use this confidence to rank the explanations to minimize the number of explanations a user must interact with before reaching a valid answer.

A weak label matrix, $\lambda$, is formed as a $e \times m$ matrix, where $m$ is the number of labeling functions, and $e$ is the total number of explanations over $n$ documents: $e = \sum_{i \in n}|e_i|$ where $e_i$ is the set of explanations for a document $i$. 
Let $\Sigma$ be the $m \times m$ covariance matrix of $\lambda$, then the parameter $\hat{\mathbf{z}}$ can be estimated by solving the following matrix completion problem:
$$
\hat{\mathbf{z}}
=\underset{\mathbf{z}}{\operatorname{argmin}}
\left\|
\left(
\boldsymbol{\Sigma}^{-1}
+
\mathbf{z z}^T\right
) 
\odot \Omega
\right\|_F
$$
where $\|\cdot\|_F$ is the Frobenius norm, and $\Omega \in \mathbb{R}^{m \times m}$ is a positive semidefinite matrix which encodes conditional independence structure amongst the labeling functions.
If $\Omega$ is correctly specified, $\hat{\mathbf{z}}_i$ will represent how often the labeling function $f_i(.)$ independently reaches the same conclusion as the other labeling functions, and with a sufficient number of labeling functions, will serve as a proxy for labeling function accuracy. The generative model is then a function that transforms $\hat{\mathbf{z}}$, and values of the labeling functions for an explanation $\lambda_{i, e=j}$ into a probabilistic label: 
$$
\hat{p}\left(y \mid \lambda_{i, e=j}\right)=f\left(\hat{\mathbf{z}}, \lambda_{i, e=j}\right)
$$

\subsection{Biegel et al.}
\label{app_sec:biegel}

As validation progresses \texttt{(explanation, value, $\lambda_{i}$, valid)} tuples are collected. We use the human-disagreement penalty proposed in \citep{weasul} to penalize the generative model when there is disagreement between the model and the validated data.

In their paper, they add a penalty to the matrix completion when optimizing for $\hat{\mathbf{z}}$. This penalty is a simple quadratic difference between the human and the probabilistic label: $\sum_{i \in D}\left(f\left(\hat{\mathbf{z}}, \lambda_{i, e=j}\right)-\mathbf{y}_{i}\right)^2$. In effect, the optimization is penalized for disagreeing with the human. The penalty is scaled by a hyper-parameter $\alpha$:
$$
\hat{p}\left(y \mid \lambda_{i, e=j}\right)=f\left(\hat{\mathbf{z}}, \lambda_{i, e=j}\right) +\alpha P e(\mathbf{z})
$$

We use $\alpha = 100$ as the authors do in their experiments.

\end{document}